\documentclass[runningheads]{llncs}
\usepackage[T1]{fontenc}
\usepackage{graphicx}
\usepackage{multirow}
\usepackage{adjustbox}
\usepackage{tabularx}
\usepackage{booktabs}
\usepackage{amsfonts}
\usepackage{pifont}

\usepackage{array}
\newcommand{\PreserveBackslash}[1]{\let\temp=\\#1\let\\=\temp}
\newcolumntype{C}[1]{>{\PreserveBackslash\centering}p{#1}}
\newcolumntype{R}[1]{>{\PreserveBackslash\raggedleft}p{#1}}
\newcolumntype{L}[1]{>{\PreserveBackslash\raggedright}p{#1}}

\begin{document}
\title{Multi-Site Class-Incremental Learning with Weighted Experts in Echocardiography}
\titlerunning{Multi-Site Class-Incremental Learning with Weighted Experts}

\author{Kit M. Bransby\inst{1,2}, Woo-Jin Cho Kim\inst{1}, Jorge Oliveira\inst{1}, Alex Thorley\inst{1}, Arian Beqiri\inst{1}, Alberto Gomez\inst{1}, Agisilaos Chartsias\inst{1}}
\authorrunning{Bransby et al.}

\institute{
Ultromics Ltd., Oxford, United Kingdom
\\ \email{agis.chartsias@ultromics.com} 
\and
Queen Mary University of London, United Kingdom
\\ \email{k.m.bransby@qmul.ac.uk} 
}

\maketitle           

\begin{abstract}
Building an echocardiography view classifier that maintains performance in real-life cases requires diverse multi-site data, and frequent updates with newly available data to mitigate model drift. Simply fine-tuning on new datasets results in ``catastrophic forgetting'', and cannot adapt to variations of view labels between sites. Alternatively, collecting all data on a single server and re-training may not be feasible as data sharing agreements may restrict image transfer, or datasets may only become available at different times. Furthermore, time and cost associated with re-training grows with every new dataset. We propose a class-incremental learning method which learns an expert network for each dataset, and combines all expert networks with a score fusion model. The influence of ``unqualified experts'' is minimised by weighting each contribution with a learnt in-distribution score. These weights promote transparency as the contribution of each expert is known during inference. Instead of using the original images, we use learned features from each dataset, which are easier to share and raise fewer licensing and privacy concerns. We validate our work on six datasets from multiple sites, demonstrating significant reductions in training time while improving view classification performance.  

\end{abstract}

\section{Introduction}\label{intro}

\indent Echocardiography (echo) view classification is often a necessary first step in automated image interpretation and analysis, as different tasks may require different views as input~\cite{vaseli2019designing}. Several deep learning methods have been proposed to this end \cite{vaseli2019designing,wegner2022accuracy,kusunose2020clinically,madani2018fast} demonstrating excellent performance. To generalise and maintain performance in real-life cases, view classifiers need to be trained on a diverse multi-site dataset to accommodate varying acquisition, demographic, and clinical parameters. However, these parameters may change over time, e.g. with machine upgrades, modification of protocols, or changes in demographics. Such changes require updating the classifier with newly available data to mitigate model drift \cite{modeldrift}. Naive approaches like fine-tuning on a new dataset may result in ``catastrophic forgetting''~\cite{catastrophic}, in which previous knowledge is forgotten at the expense of the new information. Retaining previous knowledge while learning on new data is the goal of incremental-learning~\cite{van2022three}. \\
\indent Furthermore, newly available data (e.g. from sites with different protocols) may have different sets of view labels. Labels can be characterised as ``base'', ``novel'', or ``overlapping'', depending on whether they are present in the original data, the new data, or both. The goal of class-incremental learning (CIL)~\cite{van2022three} is to learn to classify the increasing label set over time. A straightforward solution is to retrain using all data combined. However, this is not always feasible as data sharing agreements may limit access or transfer outside the acquisition site; or different datasets may be available at different times. In addition, retraining a classifier when a new dataset becomes available is inefficient as the total training time and cost grows considerably with the number and size of datasets.\\
\indent A common approach for CIL is to use a single model, and update parts or all of the weights using knowledge distillation \cite{icarl,lwf,podnet} or weight regularisation \cite{ewc,memoryawaresynapses}. Others learn additional parameters with dynamic architecture changes \cite{tao2020topology,der,foster}, or by duplicating the model for novel data and pruning \cite{yoon2017lifelong}. Wu et al. \cite{wu2022class} expand on this by duplicating and fine-tuning a base model for each new dataset, resulting in set of expert model branches, each specialised on a single dataset. They combine the output logits of the independently trained branches using a learnt score fusion (SF) network that enables knowledge transfer between base and novel classes.
This leverages shared information between representations in different branches, but is sensitive to one branch contributing ineffective information to another branch. We term this the ``unqualified expert'' problem and may happen when there are differences in data distributions, caused for instance by different label sets, patient demographics or scanner manufacturers. As a result view predictions for out-of-distribution (o.o.d.) images can be incorrect, but also highly confident. 
\\
\begin{figure}[t]
    \centering
    \caption{Examples of different views from WASE, CAMUS, Medstar, StG datasets.}
    \includegraphics[width=1\linewidth]{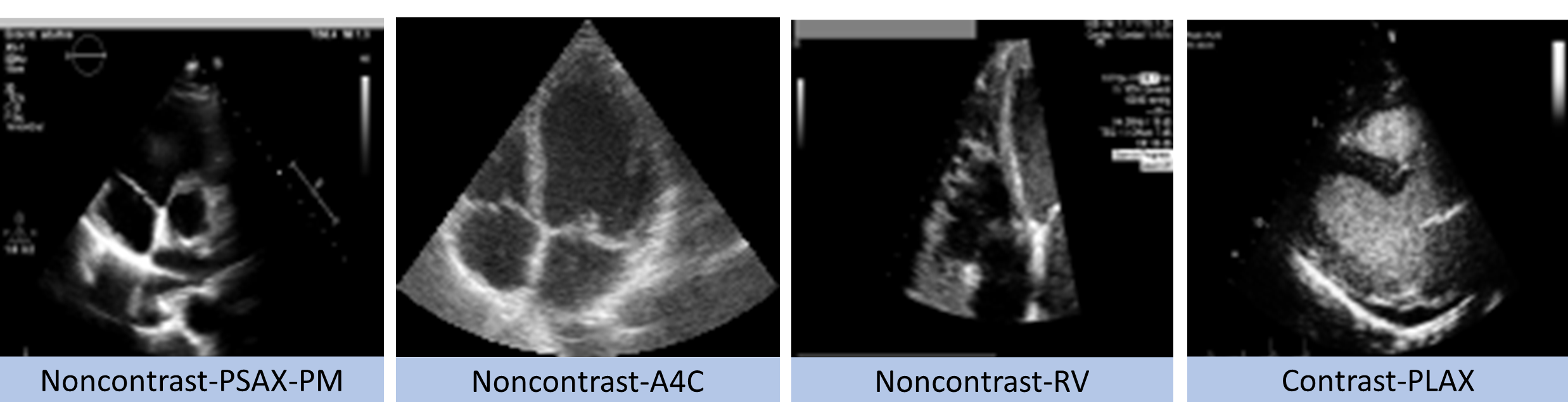}
    \label{examples}
\end{figure}
\indent In this paper we address the problem of building an echo view classifier model with multiple datasets from different sites (see Fig.~\ref{examples}) that are not simultaneously available and have different, overlapping sets of labels. To this end, we build upon~\cite{wu2022class}, and
mitigate the ``unqualified expert'' problem by weighing the contribution of each branch by a learnt in-distribution (i.d.) score. This minimises the influence of logits which are o.o.d. for a given input image, while maximising the i.d. logits. Our weighting improves transparency as the contribution of each model is known during inference, indicating the similarity between input data and different training sets. Our method has the benefit of using learnt features as input rather than images, which provides a simple workaround in cases where licensing prevents image sharing but not sharing of byproduct features. \\
\indent \textbf{\textit{Contributions:}} We apply CIL to echo view classification and propose a new i.d. score weighting that improves performance by minimising the influence of ``unqualified experts'' and promotes transparency. We validate our method on six datasets from multiple sites, demonstrating improved view classification without needing to transfer images out of remote servers. By bypassing the need for re-training with all data, the cumulative training time and cost is reduced with further reductions achieved through more incremental steps.

\section{Method}

\subsection{Problem Setting}

Given a dataset \( D = \{(x_i, y_i)\}_{i=1}^N \), where \( x_i \) represents an ultrasound image and \( y_i \) denotes the corresponding ground truth view label, the objective of view classification is to learn an image encoder $\phi$ and linear classifier \( W \in \mathbb{R}^{k \times |\mathcal{Y}|} \), where \( \mathcal{Y} \) is the set of labels in \( D \) and $k$ is the size of the image features. The view label is learnt by minimising the cross entropy loss between \( y \) and the label prediction $\hat{y} = \sigma(\phi(x)W)$ where $\sigma(\cdot)$ is a softmax function. \\
\indent We first train a base model $\mathcal{M}_{b} = \{\phi_{b},W_{bb}\} $ on a large dataset $\mathcal{D}_{b}$ with label set $\mathcal{Y}_{b}$ of base classes. We consider $t$ incremental steps $n \in \{n1, n2,..,nt\}$ each with a dataset $\mathcal{D}_{n}$ and label set $\mathcal{Y}_{n}$. There can be varying degrees of intersection between base classes $\mathcal{Y}_{b}$ and the new label set $\mathcal{Y}_{n}$. For instance, there may be novel classes not present in the base classes (\(\mathcal{Y}_{n} \cap \mathcal{Y}_{b} = \emptyset \)) or overlapping classes (\(\mathcal{Y}_{b} \cap \mathcal{Y}_{n} \neq \emptyset \)). Our goal is to accurately predict all classes, regardless of whether they are introduced in an incremental step, are novel, base, or overlapping. \\
\indent Simply combining base and incremental datasets into a single set and retraining $\mathcal{M}_{b}$ from scratch is not ideal due to cost, time, and restrictive data licensing as motivated in Section~\ref{intro}. Similarly, naively fine-tuning $\mathcal{M}_{b}$ on successive incremental datasets $\mathcal{D}_{n1}, \mathcal{D}_{n2}, .., \mathcal{D}_{nt}$ is difficult due to the changing size of the label set, and ultimately results in catastrophic forgetting of the base classes $\mathcal{Y}_{b}$. 

\begin{figure}[t!]
    \centering
    \caption{Network architecture: predictions from expert branches are re-weighted by an in-distribution score to minimise the influence of unqualified experts.}
    \includegraphics[width=1\linewidth]{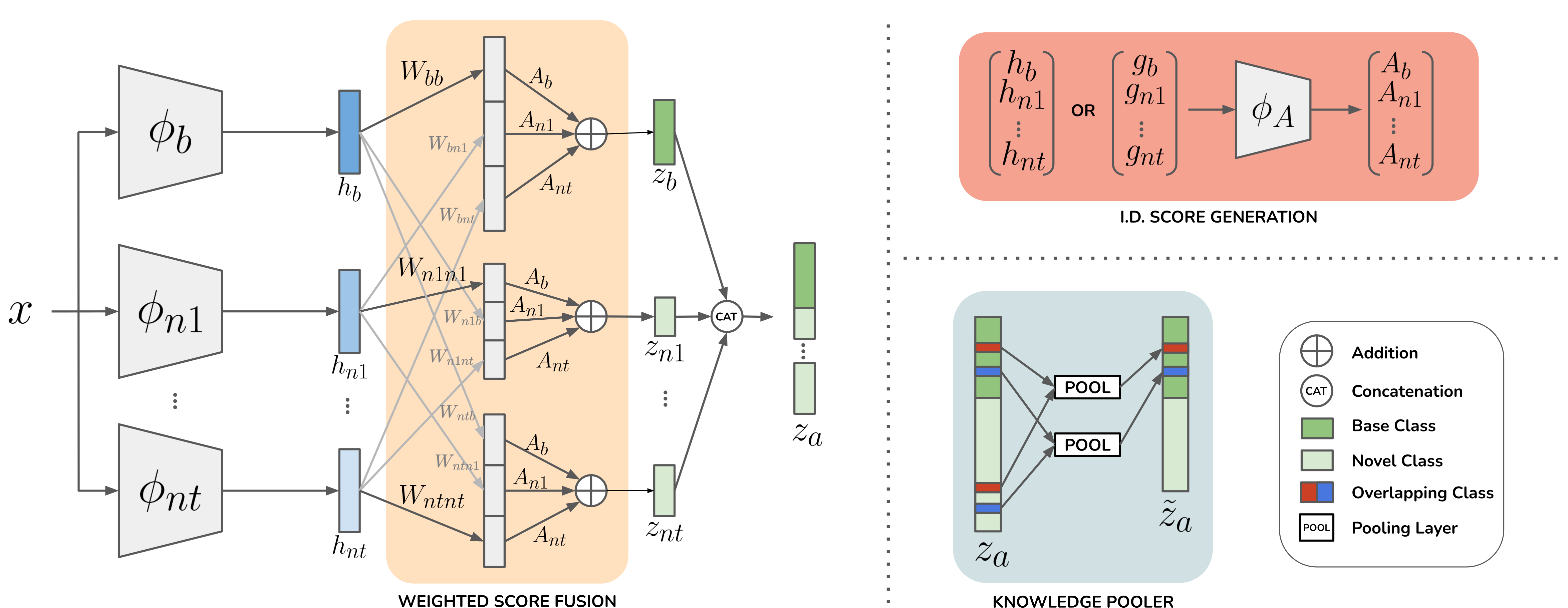}
    \label{architecture}
\end{figure}

\subsection{Score Fusion}

We follow Wu et al.~\cite{wu2022class} where $\mathcal{M}_{b}$ is cloned and fine-tuned for each incremental dataset $\mathcal{D}_{n}$, yielding $t+1$ expert models  $\{\mathcal{M}_{b}, \mathcal{M}_{n1},.. , \mathcal{M}_{nt}\}$~\footnote{We unfreeze all layers instead of freezing the first two convolutional blocks as in \cite{wu2022class}, which we find improves performance.}. This approach mitigates the problem of forgetting by retaining all weights; however, it introduces a new challenge of selecting or combining experts during inference. \\
\indent This is addressed with SF~\cite{wu2022class} that allows an expert to make a logit prediction on its label set as well as the label sets of other experts. This allows for knowledge sharing between branches and is useful for overlapping classes or shared semantics. Specifically, the weights of an expert $\{\phi_{d}, W_{dd}\}$, where $d \in \{b, n1,.., nt\}$, are frozen and used to compute a logit score  $z_{dd} = h_{d}W_{dd}$ for its label set $\mathcal{Y}_{d}$ using feature vector $h_{d} \in \mathbb{R}^{k}$ and the weight $W_{dd}$. To compute logit predictions on label sets of other experts $\mathcal{Y}_{d'}$, each expert learns an additional set of weights $W_{dd'}$ for all $d' \in \{b, n1, .., nt\}, d \neq d' $. On each branch, we compute the final logits $z_{d}$ by summing the logits contributed by each expert as follows: 

\begin{equation} \label{zd_eq}
    z_{d} = \sum_{d'=b, n1,..,nt}^{} h_{d}W_{dd'}
\end{equation} 

The logits from each branch are concatenated to give $z_{a}$. As overlapping classes may be present between incremental steps, a logit for the same class may appear several times in $z_{a}$. We follow \cite{wu2022class} with a ``knowledge pooler'' to obtain final logits $\tilde{z}_{a}$ by maxpooling the overlapping classes (See Fig~\ref{architecture}). 
This enables information sharing between experts, however some experts may have significant differences in knowledge and contribute unhelpful information that deteriorates performance. In the next section, we address the ``unqualified expert'' problem.

\subsection{Weighted Score Fusion}

Given an input image, we compute an i.d. score to weigh each expert's logits such that the influence of unqualified o.o.d experts is minimised and the influence of i.d. experts is maximised. We do so by learning an attention-like  weighting vector $A \in \mathbb{R}^{\left | d \right |}$ using a feed-forward network $\phi_{A}$ consisting of 3 linear layers, the first two with ReLU activation and the last with a softmax. \\
\indent We propose two strategies to learn $A$: (1) attention-weighted score fusion (attn-wSF) where the concatenated feature vectors of all experts $h \in \mathbb{R}^{k \left | d \right |}$ are used as input to $\phi_{A}$; and (2) neural mean discrepancy weighted score fusion (nmd-wSF), which uses the neural mean discrepancy (nmd) \cite{nmd} as input. Nmd is an out-of-distribution metric which measures the discrepancy between neural means of an input image and the average neural mean of the i.d. training set. The neural mean is calculated by averaging output activations of the convolutional layers. Like with attn-wSF, we compute a nmd vector $g_{d}$ for each expert and concatenate them to $g \in \mathbb{R}^{p \left | d \right |}$ where $p$ is the size of the nmd vector. \\
\noindent Similar to Equation \ref{zd_eq}, we compute the final logits for each branch as follows:
\begin{equation}
    z_{d} = \sum_{d'=b, n1,..,nt}^{} h_{d}W_{dd'}A_{d}
\end{equation}

\subsection{Multi-Site Training}

Assuming that the incremental datasets are acquired at different medical sites, which often impose data sharing restrictions, we propose a multi-site training paradigm, which does not require transferring the image data. For every step, the view classifier is cloned, transferred into the remote server, and fine-tuned on dataset $\mathcal{D}_{n}$. We then make five forward passes of the fine-tuned classifier $\mathcal{M}_{n}$ on randomly augmented data from $\mathcal{D}_{n}$ to generate five different $g$ and $h$ vectors for each example. We then transfer these vectors back to the local server, which are used as input to train the weighted score fusion network.

\begin{table}[t!]
\caption{Frequency of view labels in training, validation and test sets. }
\begin{adjustbox}{width=1\columnwidth}
\begin{tabular}{L{1.55cm}C{1.4cm}|C{1cm}C{1cm}C{1cm}C{1cm}|C{1cm}C{1cm}C{1cm}C{1cm}|C{1cm}C{1cm}C{1cm}C{1cm}|C{1cm}C{1cm}}
\toprule
\multirow{2}{*}{View} & \multirow{2}{*}{Contrast} & \multicolumn{4}{c|}{Train} & \multicolumn{4}{c|}{Validation} & \multicolumn{4}{c|}{Internal Test} & \multicolumn{2}{c}{External Test} \\
&  & Wase & Camus & Mstr & StG & Wase & Camus & Mstr & StG & Wase & Camus & Mstr & StG & Mahi & UoC  \\
 \midrule
A2C & \ding{51} & 0 & 0 & 0 & 1182 & 0 & 0 & 0 & 132 & 0 & 0 & 0 & 116 & 845 & 36 \\
A3C & \ding{51} & 0 & 0 & 0 & 1208 & 0 & 0 & 0 & 137 & 0 & 0 & 0 & 119 & 429 & 23 \\
A4C & \ding{51} & 0 & 0 & 0 & 1218 & 0 & 0 & 0 & 138 & 0 & 0 & 0 & 119 & 726 & 58 \\
PLAX & \ding{51} & 0 & 0 & 0 & 923 & 0 & 0 & 0 & 125 & 0 & 0 & 0 & 86 & 6 & 4 \\
A2C & \ding{55} & 4361 & 407 & 886 & 0 & 559 & 79 & 87 & 0 & 559 & 48 & 136 & 0 & 716 & 944 \\
A3C & \ding{55} & 4012 & 0 & 900 & 0 & 501 & 0 & 87 & 0 & 492 & 0 & 108 & 0 & 679 & 493 \\
A4C & \ding{55} & 6474 & 407 & 1668 & 0 & 811 & 42 & 166 & 0 & 801 & 48 & 202 & 0 & 825 & 1806 \\
A5C & \ding{55} & 1127 & 0 & 355 & 0 & 127 & 0 & 40 & 0 & 138 & 0 & 44 & 0 & 147 & 145 \\
PLAX & \ding{55} & 4661 & 0 & 1176 & 0 & 614 & 0 & 125 & 0 & 566 & 0 & 138 & 0 & 929 & 1072 \\
PLAX-AV & \ding{55} & 1480 & 0 & 223 & 0 & 190 & 0 & 24 & 0 & 185 & 0 & 23 & 0 & 306 & 113 \\
PSAX-AV & \ding{55} & 3936 & 0 & 448 & 0 & 488 & 0 & 44 & 0 & 515 & 0 & 60 & 0 & 597 & 612 \\
PSAX-PM & \ding{55} & 3351 & 0 & 755 & 0 & 420 & 0 & 93 & 0 & 435 & 0 & 80 & 0 & 310 & 543 \\
RV & \ding{55} & 0 & 0 & 462 & 0 & 0 & 0 & 72 & 0 & 0 & 0 & 60 & 0 & 60 & 318 \\
SC & \ding{55} & 0 & 0 & 589 & 0 & 0 & 0 & 72 & 0 & 0 & 0 & 62 & 0 & 482 & 328 \\
SC-IVC & \ding{55} & 0 & 0 & 398 & 0 & 0 & 0 & 47 & 0 & 0 & 0 & 49 & 0 & 242 & 275 \\ \midrule
\textbf{Total}  & & 29402 & 814 & 7860 & 4531 & 3710 & 84 & 849 & 532 & 3691 & 96 & 962 & 440 & 7299 & 6700 \\ \bottomrule
\end{tabular}
\end{adjustbox}
\label{labels}
\end{table}

\section{Experiments \& Results}

\subsection{Datasets}

As base dataset $\mathcal{D}_{b}$, we use WASE-Normals (WASE)~\cite{wase}, a large multi-site proprietary dataset of 2,009 healthy volunteers acquired at 18 sites from 15 countries. We also use three incremental datasets: (1) CAMUS~\cite{camus}, a public dataset acquired from 500 patients at University Hospital of St Etienne, half of which are considered at pathological risk; (2) Medstar WASE-Covid (Medstar)~\cite{mstr}, a multi-site proprietary database of 870 patients with COVID-19 at 13 sites from 9 countries; 3) St George's (StG), a proprietary stress echo dataset of 420 patients some with coronary artery disease. We validate our method on an internal test set sampled from the four datasets and an independent external test set from two proprietary datasets: (1) Mazankowski Alberta Heart Institute (MAHI) consisting of 250 patients undergoing chemotherapy with cardiotoxic drugs; (2) University of Chicago (UoC) consisting of 391 patients with cardiomyopathy. \\
\indent We extract images from 15 echo view labels
and artificially decrease the overlap between datasets by removing non-contrast examples from StG and contrast examples from Medstar. As shown in Table~\ref{labels}, the resulting datasets have different label sets with varying degrees of overlap.
Each dataset is split into train (80\%), validation (10\%) and test (10\%) sets at a patient level. A single random frame (112$\times$112) was sampled from each echo video. The final dataset has 42,622 train, 5,175 validation, 5,189 internal test, and 13,999 external test frames. 

\subsection{Implementation \& Training}

View classifiers use ResNet18~\cite{resnet} architecture and were trained for 200 epochs on a NVIDIA GeForce RTX 2080Ti with cross-entropy loss, Adam optimiser, batch size of 64, and learning rate of 1e-3. The score fusion networks were trained for 50 epochs with the same settings. Weights from the epoch with the highest validation accuracy were saved. Hyperparameters were tuned on a held-out validation set. We evaluate classification performance using accuracy and F1-score metrics. Our method is implemented in PyTorch, and the code is available here: https://github.com/kitbransby/class-incremental-learning-echo

\begin{table}[t!]
\centering
\caption{Quantitative Results for Internal Test Set (WASE, CAMUS, Medstar, StG datasets). Best results in bold, and second best underlined}
\begin{adjustbox}{width=1\columnwidth}
\begin{tabular}{L{4cm}|C{1.6cm}|C{2cm}|C{0.9cm}C{0.9cm}|C{0.9cm}C{0.9cm}|C{0.9cm}C{0.9cm}|C{0.9cm}C{0.9cm}|C{0.9cm}C{0.9cm}}
\toprule
\multirow{2}{*}{Experiment} & \multirow{2}{*}{\#Experts} & \multirow{2}{*}{Data Transfer} & \multicolumn{2}{c|}{WASE} & \multicolumn{2}{c|}{CAMUS} & \multicolumn{2}{c|}{Medstar} & \multicolumn{2}{c|}{StG} & \multicolumn{2}{c}{Average} \\
 & &  & Acc & F1 & Acc & F1 & Acc & F1 & Acc & F1 & Acc & F1 \\
\midrule
Fine-Tuning (Constant) & 1 & None & 0.0 & 0.0 & 0.0 & 0.0 & 0.0 & 0.0 & 83.9 & 83.7 & 7.0 & 3.9 \\
Fine-Tuning (Expand) & 1 & None & 1.9 & 3.1 & 0.0 & 0.0 & 3.9 & 4.3 & 85.5 & 87.2 & 9.3 & 4.8 \\
Single Expert (WASE)  & 1 & None & \textbf{94.5} & \textbf{94.5} & 76.0 & 80.9 & 65.1 & 58.7 & 0.0 & 0.0 & 68.6 & 60.2 \\
Single Expert (CAMUS)  & 1 & None & 24.4 & 15.8 & \textbf{95.8} & \textbf{95.8} & 25.3 & 16.1 & 0.0 & 0.0 & 22.6 & 13.8 \\
Single Expert (Medstar)  & 1 & None & 80.6 & 83.0 & 69.8 & 76.2 & 80.0 & 79.5 & 0.0 & 0.0 & 74.2 & 72.0 \\
Single Expert (StG) & 1 & None & 0.0 & 0.0 & 0.0 & 0.0 & 0.0 & 0.0 & \textbf{94.1} & \textbf{94.1} & 6.6 & 3.4 \\
\midrule
Max Logit \cite{maxlogit} & 4 & None & 62.5 & 71.3 & 84.4 & 89.4 & 63.1 & 68.4 & 71.4 & 81.0 & 63.8 & 69.3 \\
MSP \cite{msp} & 4 & None & 56.1 & 61.9 & 92.7 & 94.2 & 45.0 & 49.2 & 59.8 & 72.0 & 55.0 & 58.6 \\
Confidence Routing \cite{wu2022class} & 4 & None & 63.4 & 68.6 & 92.7 & 94.2 & 46.7 & 50.8 & 51.8 & 66.0 & 59.9 & 63.4 \\
\midrule
SF \cite{wu2022class} & 4 & Features & 93.9 & 94.0 & 89.6 & 90.4 & 80.3 & 80.0 & 92.7 & 93.6 & 91.2 & 91.1 \\
attn-wSF (ours) & 4 & Features & \underline{94.1} & 94.3 & \textbf{95.8} & \textbf{95.8} & \underline{81.4} & \underline{80.9} & \underline{93.9} & \underline{94.0} & \underline{91.8} & \underline{91.7} \\
nmd-wSF (ours) & 4 & Features & \textbf{94.5} & \underline{94.4} & \underline{94.8} & \underline{95.3} & 80.8 & 80.5 & \textbf{94.1} & \textbf{94.1} & \textbf{91.9} & \textbf{91.9} \\
\midrule
Combine \& Retrain (oracle) & 1 & Images & \underline{94.1} & 94.2 & \textbf{95.8} & \textbf{95.8} & \textbf{82.1} & \textbf{82.1} & 90.2 & 90.5 & 91.6 & 91.5 \\
\bottomrule
\end{tabular}
\end{adjustbox}
\label{internal_quant}
\end{table}

\begin{table}[h!]
\centering
\caption{Quantitative Results for External Test Set (MAHI, UoC datasets). Best results in bold, and second best underlined}
\begin{adjustbox}{width=0.85\columnwidth}
\begin{tabular}{L{4.1cm}|C{1.8cm}|C{2cm}|C{1cm}C{1cm}|C{1cm}C{1cm}|C{1cm}C{1cm}}
\toprule
\multirow{2}{*}{Experiment} & \multirow{2}{*}{\# Experts} & \multirow{2}{*}{Data Transfer} & \multicolumn{2}{c|}{MAHI} & \multicolumn{2}{c|}{UoC} & \multicolumn{2}{c}{Average} \\
& &  & Acc & F1 & Acc & F1 & Acc & F1 \\
\midrule
Fine-Tuning (Constant) & 1  & None & 12.8 & 10.3 & 0.7 & 0.1 & 7.0  & 3.9 \\
Fine-Tuning (Expand) & 1  & None & 17.1 & 13.3 & 3.8 & 4.9 & 10.7 & 8.2 \\
Single Expert (WASE) & 1 & None & 55.9 & 45.5 & 72.9 & 67.8 & 64.1 & 55.1 \\
Single Expert (CAMUS) & 1  & None & 13.1 & 6.6 & 32.0 & 20.4 & 22.2 & 13.4 \\
Single Expert (Medstar) & 1  & None & 56.9 & 51.4 & 79.0 & 79.1 & 69.0 & 64.1 \\
Single Expert (StG) & 1  & None & 11.2 & 8.0 & 0.1 & 0.0 & 6.1 & 3.6 \\
\midrule
Max Logit \cite{maxlogit} & 4 & None & 43.2 & 44.0 & 61.2 & 68.6 & 51.8 & 54.7 \\
MSP \cite{msp} & 4 & None & 34.4 & 35.4 & 55.4 & 61.6 & 44.5 & 46.9 \\
Confidence Routing \cite{wu2022class} & 4 & None & 38.2 & 39.4 & 57.1 & 63.2 & 47.3 & 49.7 \\
\midrule
SF \cite{wu2022class} & 4 & Features & \underline{72.0} & 68.3 & 81.6 & 80.8 & 76.6 & 74.2 \\
attn-wSF (ours) & 4 & Features & \textbf{72.5} & \textbf{69.6} & \underline{82.6} & \underline{82.1} & \textbf{77.4} & \textbf{75.6} \\
nmd-wSF (ours) & 4 & Features & 71.6 & \underline{69.5} & 79.8 & 79.5 & 75.5 & 74.3 \\
\midrule
Combine \& Retrain (oracle) & 1  & Images & 71.4 & 69.0 & \textbf{82.7} & \textbf{82.7} & \underline{76.8} & \underline{75.1} \\
\bottomrule
\end{tabular}
\end{adjustbox}
\label{external_quant}
\end{table}

\subsection{Comparison to Existing Methods \& Ablation Study}

We validate our method on an internal test set (WASE, CAMUS, Medstar, StG) to measure performance on incremental datasets seen during training, and an external set (MAHI, UoC) of data not seen during training to measure generalisability. Results are presented in Tables~\ref{internal_quant} and~\ref{external_quant}, respectively. \\
\indent We demonstrate that naively fine-tuning a classifier on successive datasets results in ``catastrophic forgetting''. We test two configurations, one where the classifier head is replaced at every step, and a second that expands to accommodate novel classes~\cite{lomonaco2017core50}. These do not require data transfers and perform well on the final dataset (StG) but forget learning from previous data. We also explore other methods that do not require data transfer such as using a single expert and find they perform well on i.d. data but do not generalise well on o.o.d data. \\
\begin{figure}[t!]
    \centering
    \caption{Distribution of attention scores across a selection of test sets using attn-wSF (top row) and nmd-wSF (bottom row). Note when using nmd, the attention scores are pushed towards 0 and 1 which may reduce generalisability }
    \includegraphics[width=1\linewidth]{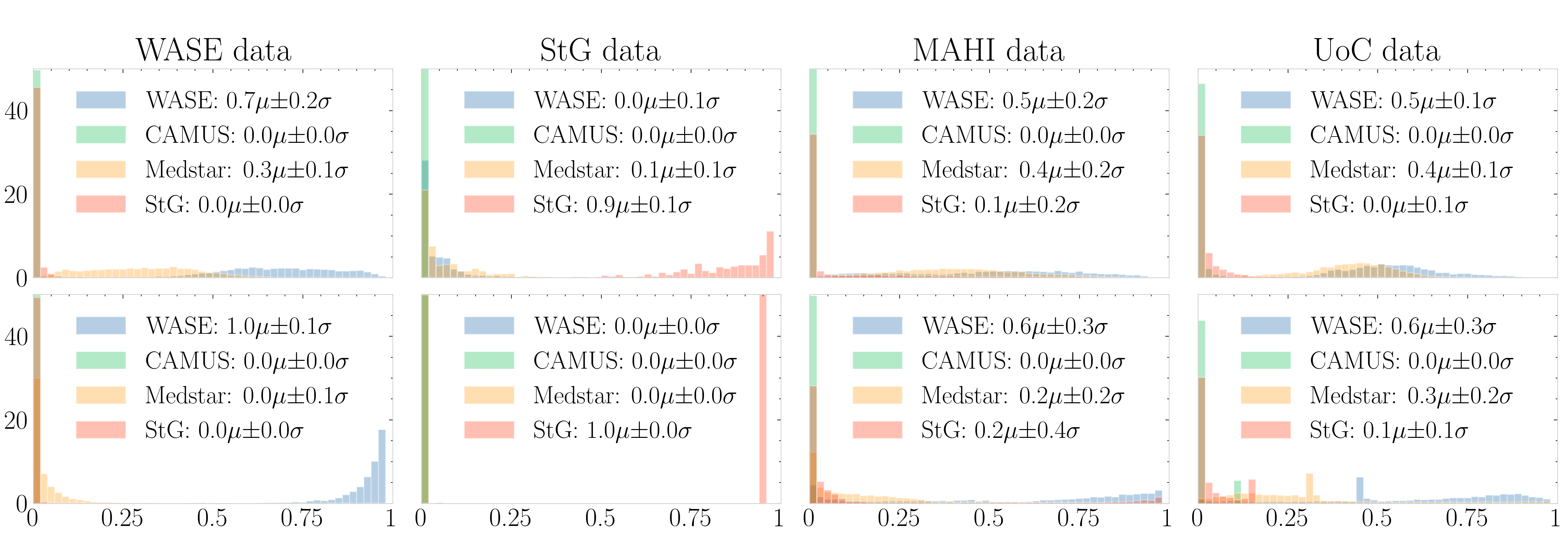}
    \label{attn}
\end{figure}
\indent Our primary comparison is to the training oracle where the model is retrained with all data combined at each incremental step. This method is preferred given unlimited compute, time constraints and data access, which are not typically available. On average our method outperforms the oracle on both the internal and external sets, including four out of six individual datasets. \\
\indent In addition, we ablate our weighting method, demonstrating that attn-wSF and nmd-wSF lead to an improvement in performance when compared to SF \cite{wu2022class}. We find that nmd-wSF performs best on the internal test set, however, attn-wSF generalises better to the external test set. We theorise that nmd vectors for each dataset are quite distinct, therefore the i.d. scores are pushed towards 0 and 1 during training (see Fig. \ref{attn}). When presented with datasets from the external test set the attention mechanism deteriorates as it considers the input image o.o.d. for all experts. \\ 
\indent Finally, we compare different confidence-based methods that select the best prediction from multiple experts, these include: (1) Max Logit~\cite{maxlogit} that selects the max logit for each class; (2) MSP~\cite{msp}, that selects the max softmax for each class; (3) Confidence Routing~\cite{wu2022class} that selects the expert with the highest softmax probability, and uses its softmax prediction as the final prediction. All such approaches perform poorly, perhaps as classifiers are optimised to be overconfident leading to predictions which are uncorrelated with confidence measures. \\
\begin{figure}[t!]
    \centering
    \caption{Efficiency analysis: Cumulative training time (a) and inference time (b). }
    \includegraphics[width=0.8\linewidth]{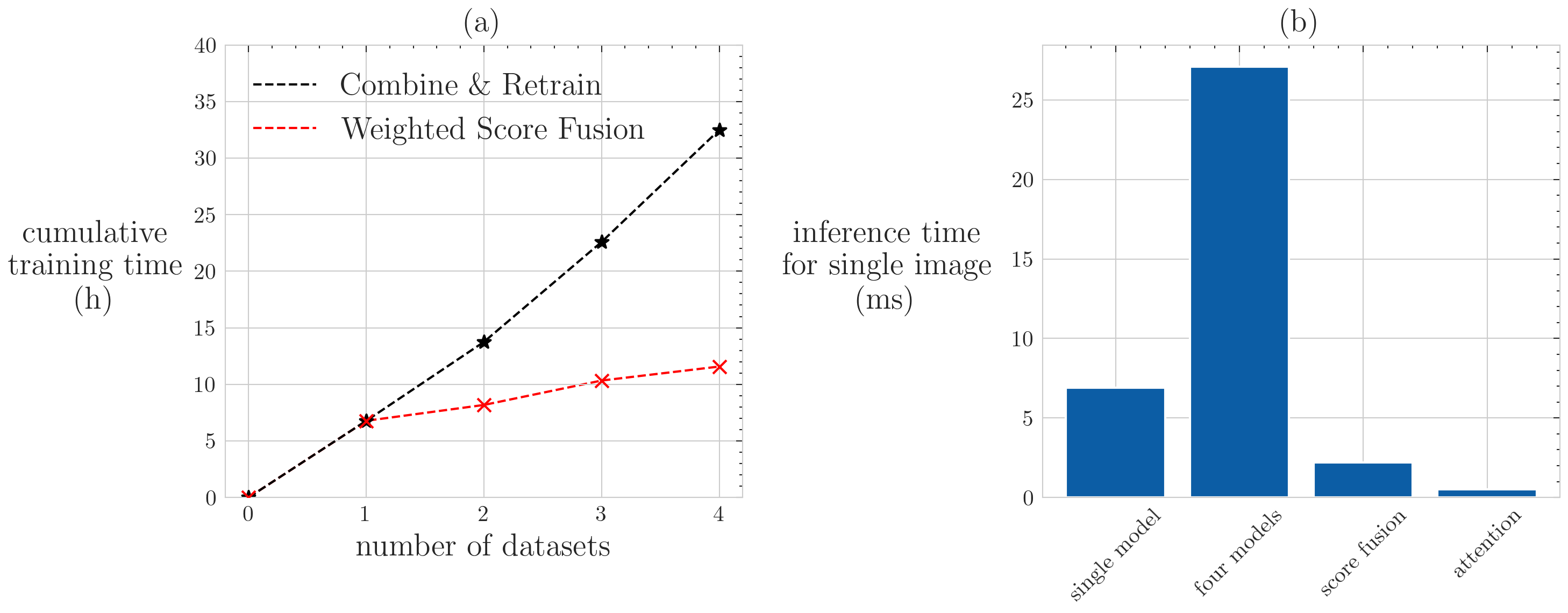}
    \label{efficiency}
\end{figure}
\indent Fig.~\ref{efficiency} shows that our wSF methods reduce the cumulative training time, and this grows significantly with the number of datasets. Our attention-based weighting module has minimal computational overhead. While inference time is larger when using multiple experts, this can be reduced with parallelisation. 

\section{Conclusion}

Echo view classifiers require training on diverse multi-site data, and frequent updates with newly acquired data to avoid model drift. We address the problem of updating such a model using multiple datasets with partly overlapping label sets that are from different sites and may not be simultaneously available. We propose attn-wSF and nmd-wSF, two CIL methods that train and combine predictions from multiple experts models with a learnt i.d. weighting which minimises the influence of o.o.d predictions from ``unqualified experts''. Our results demonstrate improved view classification with significantly reduced cumulative training time and without needing to transfer images out of remote servers. We aim to extend our wSF methods by training multiple experts within a single branch, which would reduce the memory requirements and inference time. 



\bibliographystyle{splncs04}
\bibliography{ref}

\begin{thebibliography}{10}
\providecommand{\url}[1]{\texttt{#1}}
\providecommand{\urlprefix}{URL }
\providecommand{\doi}[1]{https://doi.org/#1}

\bibitem{memoryawaresynapses}
Aljundi, R., Babiloni, F., Elhoseiny, M., Rohrbach, M., Tuytelaars, T.: Memory aware synapses: Learning what (not) to forget. In: Proceedings of the European conference on computer vision (ECCV). pp. 139--154 (2018)

\bibitem{wase}
Asch, F.M., Banchs, J., Price, R., Rigolin, V., Thomas, J.D., Weissman, N.J., Lang, R.M.: Need for a global definition of normative echo values—rationale and design of the world alliance of societies of echocardiography normal values study (wase). Journal of the American Society of Echocardiography  \textbf{32}(1),  157--162 (2019)

\bibitem{nmd}
Dong, X., Guo, J., Li, A., Ting, W.T., Liu, C., Kung, H.: Neural mean discrepancy for efficient out-of-distribution detection. In: Proceedings of the IEEE/CVF Conference on Computer Vision and Pattern Recognition. pp. 19217--19227 (2022)

\bibitem{podnet}
Douillard, A., Cord, M., Ollion, C., Robert, T., Valle, E.: Podnet: Pooled outputs distillation for small-tasks incremental learning. In: Computer vision--ECCV 2020: 16th European conference, Glasgow, UK, August 23--28, 2020, proceedings, part XX 16. pp. 86--102. Springer (2020)

\bibitem{resnet}
He, K., Zhang, X., Ren, S., Sun, J.: Deep residual learning for image recognition. In: Proceedings of the IEEE conference on computer vision and pattern recognition. pp. 770--778 (2016)

\bibitem{maxlogit}
Hendrycks, D., Basart, S., Mazeika, M., Zou, A., Kwon, J., Mostajabi, M., Steinhardt, J., Song, D.: Scaling out-of-distribution detection for real-world settings. In: International Conference on Machine Learning. pp. 8759--8773. PMLR (2022)

\bibitem{msp}
Hendrycks, D., Gimpel, K.: A baseline for detecting misclassified and out-of-distribution examples in neural networks. In: International Conference on Learning Representations (2016)

\bibitem{mstr}
Karagodin, I., Singulane, C.C., Woodward, G.M., Xie, M., Tucay, E.S., Rodrigues, A.C.T., Vasquez-Ortiz, Z.Y., Alizadehasl, A., Monaghan, M.J., Salazar, B.A.O., et~al.: Echocardiographic correlates of in-hospital death in patients with acute covid-19 infection: the world alliance societies of echocardiography (wase-covid) study. Journal of the American Society of Echocardiography  \textbf{34}(8),  819--830 (2021)

\bibitem{ewc}
Kirkpatrick, J., Pascanu, R., Rabinowitz, N., Veness, J., Desjardins, G., Rusu, A.A., Milan, K., Quan, J., Ramalho, T., Grabska-Barwinska, A., et~al.: Overcoming catastrophic forgetting in neural networks. Proceedings of the national academy of sciences  \textbf{114}(13),  3521--3526 (2017)

\bibitem{kusunose2020clinically}
Kusunose, K., Haga, A., Inoue, M., Fukuda, D., Yamada, H., Sata, M.: Clinically feasible and accurate view classification of echocardiographic images using deep learning. Biomolecules  \textbf{10}(5), ~665 (2020)

\bibitem{camus}
Leclerc, S., Smistad, E., Pedrosa, J., {\O}stvik, A., Cervenansky, F., Espinosa, F., Espeland, T., Berg, E.A.R., Jodoin, P.M., Grenier, T., et~al.: Deep learning for segmentation using an open large-scale dataset in 2d echocardiography. IEEE transactions on medical imaging  \textbf{38}(9),  2198--2210 (2019)

\bibitem{lwf}
Li, Z., Hoiem, D.: Learning without forgetting. IEEE transactions on pattern analysis and machine intelligence  \textbf{40}(12),  2935--2947 (2017)

\bibitem{lomonaco2017core50}
Lomonaco, V., Maltoni, D.: Core50: a new dataset and benchmark for continuous object recognition. In: Conference on robot learning. pp. 17--26. PMLR (2017)

\bibitem{madani2018fast}
Madani, A., Arnaout, R., Mofrad, M., Arnaout, R.: Fast and accurate view classification of echocardiograms using deep learning. NPJ digital medicine  \textbf{1}(1), ~6 (2018)

\bibitem{catastrophic}
McCloskey, M., Cohen, N.J.: Catastrophic interference in connectionist networks: The sequential learning problem. In: Psychology of learning and motivation, vol.~24, pp. 109--165. Elsevier (1989)

\bibitem{icarl}
Rebuffi, S.A., Kolesnikov, A., Sperl, G., Lampert, C.H.: icarl: Incremental classifier and representation learning. In: Proceedings of the IEEE conference on Computer Vision and Pattern Recognition. pp. 2001--2010 (2017)

\bibitem{modeldrift}
Sahiner, B., Chen, W., Samala, R.K., Petrick, N.: Data drift in medical machine learning: implications and potential remedies. The British Journal of Radiology  \textbf{96}(1150),  20220878 (2023)

\bibitem{tao2020topology}
Tao, X., Chang, X., Hong, X., Wei, X., Gong, Y.: Topology-preserving class-incremental learning. In: Computer Vision--ECCV 2020: 16th European Conference, Glasgow, UK, August 23--28, 2020, Proceedings, Part XIX 16. pp. 254--270. Springer (2020)

\bibitem{vaseli2019designing}
Vaseli, H., Liao, Z., Abdi, A.H., Girgis, H., Behnami, D., Luong, C., Dezaki, F.T., Dhungel, N., Rohling, R., Gin, K., et~al.: Designing lightweight deep learning models for echocardiography view classification. In: Medical Imaging 2019: Image-Guided Procedures, Robotic Interventions, and Modeling. vol. 10951, pp. 93--99. SPIE (2019)

\bibitem{van2022three}
Van~de Ven, G.M., Tuytelaars, T., Tolias, A.S.: Three types of incremental learning. Nature Machine Intelligence  \textbf{4}(12),  1185--1197 (2022)

\bibitem{foster}
Wang, F.Y., Zhou, D.W., Ye, H.J., Zhan, D.C.: Foster: Feature boosting and compression for class-incremental learning. In: European conference on computer vision. pp. 398--414. Springer (2022)

\bibitem{wegner2022accuracy}
Wegner, F.K., Benesch~Vidal, M.L., Niehues, P., Willy, K., Radke, R.M., Garthe, P.D., Eckardt, L., Baumgartner, H., Diller, G.P., Orwat, S.: Accuracy of deep learning echocardiographic view classification in patients with congenital or structural heart disease: importance of specific datasets. Journal of Clinical Medicine  \textbf{11}(3), ~690 (2022)

\bibitem{wu2022class}
Wu, T.Y., Swaminathan, G., Li, Z., Ravichandran, A., Vasconcelos, N., Bhotika, R., Soatto, S.: Class-incremental learning with strong pre-trained models. In: Proceedings of the IEEE/CVF Conference on Computer Vision and Pattern Recognition. pp. 9601--9610 (2022)

\bibitem{der}
Yan, S., Xie, J., He, X.: Der: Dynamically expandable representation for class incremental learning. In: Proceedings of the IEEE/CVF conference on computer vision and pattern recognition. pp. 3014--3023 (2021)

\bibitem{yoon2017lifelong}
Yoon, J., Yang, E., Lee, J., Hwang, S.J.: Lifelong learning with dynamically expandable networks. In: International Conference on Learning Representations (ICLR) (2017)

\end{thebibliography}

\end{document}